# A Parallel Framework for Multilayer Perceptron for Human Face Recognition

**Debotosh Bhattacharjee**                                      debotosh@indiatimes.com
*Reader,*
*Department of Computer Science and Engineering,*
*Jadavpur University,*
*Kolkata- 700032, India.*

**Mrinal Kanti Bhowmik**                                        mkb_cse@yahoo.co.in
*Lecturer,*
*Department of Computer Science and Engineering,*
*Tripura University (A Central University),*
*Suryamaninagar- 799130, Tripura, India.*

**Mita Nasipuri**                                               mitanasipuri@gmail.com
*Professor,*
*Department of Computer Science and Engineering,*
*Jadavpur University,*
*Kolkata- 700032, India.*

**Dipak Kumar Basu**                                            dipakkbasu@gmail.com
*Professor, AICTE Emeritus Fellow,*
*Department of Computer Science and Engineering,*
*Jadavpur University,*
*Kolkata- 700032, India.*

**Mahantapas Kundu**                                            mkundu@cse.jdvu.ac.in
*Professor,*
*Department of Computer Science and Engineering,*
*Jadavpur University,*
*Kolkata- 700032, India.*

## Abstract

Artificial neural networks have already shown their success in face recognition and similar complex pattern recognition tasks. However, a major disadvantage of the technique is that it is extremely slow during training for larger classes and hence not suitable for real-time complex problems such as pattern recognition. This is an attempt to develop a parallel framework for the training algorithm of a perceptron. In this paper, two general architectures for a Multilayer Perceptron (MLP) have been demonstrated. The first architecture is All-Class-in-One-Network (ACON) where all the classes are placed in a single network and the second one is One-Class-in-One-Network (OCON) where an individual single network is responsible for each and every class. Capabilities of these two architectures were compared and verified in solving human face recognition, which is a complex pattern recognition task where several factors affect the recognition performance like pose variations, facial expression changes, occlusions, and most importantly illumination changes. Both the structures were

D. Bhattacharjee,  M. K. Bhowmik, M. Nasipuri, D. K. Basu & M. Kundu

implemented and tested for face recognition purpose and experimental results show that the OCON structure performs better than the generally used ACON ones in term of training convergence speed of the network. Unlike the conventional sequential approach of training the neural networks, the OCON technique may be implemented by training all the classes of the face images simultaneously.

## 1.  INTRODUCTION

Neural networks, with their remarkable ability to derive meaning from complicated or imprecise data, can be used to extract patterns and detect trends that are too complex to be noticed by either humans or other computer techniques. A trained neural network can be thought of as an "expert" in the category of information it has been given to analyze [1]. This proposed work describes the way by which an Artificial Neural Network (ANN) can be designed and implemented over a parallel or distributed environment to reduce its training time. Generally, an ANN goes through three different steps: training of the network, testing of it and final use of it. The final structure of an ANN is generally found out experimentally. This requires huge amount of computation. Moreover, the training time of an ANN is very large, when the classes are linearly non-separable and overlapping in nature. Therefore, to save computation time and in order to achieve good response time the obvious choice is either a high-end machine or a system which is collection of machines with low computational power.

In this work, we consider multilayer perceptron (MLP) for human face recognition, which has many real time applications starting from automatic daily attendance checking, allowing the authorized people to enter into highly secured area, in detecting and preventing criminals and so on. For all these cases, response time is very critical. Face recognition has the benefit of being passive, nonintrusive system for verifying personal identity. The techniques used in the best face recognition systems may depend on the application of the system.

Human face recognition is a very complex pattern recognition problem, altogether. There is no stability in the input pattern due to different expressions, adornments in the input images. Sometimes, distinguishing features appear similar and produce a very complex situation to take decision. Also, there are several other that make the face recognition task complicated.  Some of them are given below.

a)  Background of the face image can be a complex pattern or almost same as the color of the face.
b)  Different illumination level, at different parts of the image.
c)  Direction of illumination may vary.
d)  Tilting of face.
e)  Rotation of face with different angle.
f)  Presence/absence of beard and/or moustache
g)  Presence/Absence of spectacle/glasses.
h)  Change in expressions such as disgust, sadness, happiness, fear, anger, surprise etc.
i)  Deliberate change in color of the skin and/or hair to disguise the designed system.

From above discussion it can now be claimed that the face recognition problem along with face detection, is very complex in nature. To solve it, we require some complex neural network, which takes large amount of time to finalize its structure and also to settle its parameters.
In this work, a different architecture has been used to train a multilayer perceptron in faster way. Instead of placing all the classes in a single network, individual networks are used for each of the classes. Due to lesser number of samples and conflicts in the belongingness of patterns to their respective classes, a later model appears to be faster in comparison to former.

## 2. ARTIFICIAL NEURAL NETWORK

Artificial neural networks (ANN) have been developed as generalizations of mathematical models of biological nervous systems. A first wave of interest in neural networks (also known as connectionist models or parallel distributed processing) emerged after the introduction of simplified neurons by McCulloch and Pitts (1943).The basic processing elements of neural networks are called artificial neurons, or simply neurons or nodes. In a simplified mathematical model of the neuron, the effects of the synapses are represented by connection weights that modulate the effect of the associated input signals, and the nonlinear characteristic exhibited by neurons is represented by a transfer function. The neuron impulse is then computed as the weighted sum of the input signals, transformed by the transfer function. The learning capability of an artificial neuron is achieved by adjusting the weights in accordance to the chosen learning algorithm. A neural network has to be configured such that the application of a set of inputs produces the desired set of outputs. Various methods to set the strengths of the connections exist. One way is to set the weights explicitly, using a priori knowledge. Another way is to train the neural network by feeding it teaching patterns and letting it change its weights according to some learning rule. The learning situations in neural networks may be classified into three distinct sorts. These are supervised learning, unsupervised learning, and reinforcement learning. In supervised learning, an input vector is presented at the inputs together with a set of desired responses, one for each node, at the output layer. A forward pass is done, and the errors or discrepancies between the desired and actual response for each node in the output layer are found. These are then used to determine weight changes in the net according to the prevailing learning rule. The term supervised originates from the fact that the desired signals on individual output nodes are provided by an external teacher [3]. Feed-forward networks had already been used successfully for human face recognition. Feed-forward means that there is no feedback to the input. Similar to the way that human beings learn from mistakes, neural networks also could learn from their mistakes by giving feedback to the input patterns. This kind of feedback would be used to reconstruct the input patterns and make them free from error; thus increasing the performance of the neural networks. Of course, it is very complex to construct such types of neural networks. These kinds of networks are called as auto associative neural networks. As the name implies, they use back-propagation algorithms. One of the main problems associated with back-propagation algorithms is local minima. In addition, neural networks have issues associated with learning speed, architecture selection, feature representation, modularity and scaling. Though there are problems and difficulties, the potential advantages of neural networks are vast. Pattern recognition can be done both in normal computers and neural networks. Computers use conventional arithmetic algorithms to detect whether the given pattern matches an existing one. It is a straightforward method. It will say either yes or no. It does not tolerate noisy patterns. On the other hand, neural networks can tolerate noise and, if trained properly, will respond correctly for unknown patterns. Neural networks may not perform miracles, but if constructed with the proper architecture and trained correctly with good data, they will give amazing results, not only in pattern recognition but also in other scientific and commercial applications [4].

**2A. Network Architecture**
The computing world has a lot to gain from neural networks. Their ability to learn by example makes them very flexible and powerful. Once a network is trained properly there is no need to devise an algorithm in order to perform a specific task; i.e. no need to understand the internal mechanisms of that task. The architecture of any neural networks generally used is All-Class-in-One-Network (ACON), where all the classes are lumped into one super-network. Hence, the implementation of such ACON structure in parallel environment is not possible. Also, the ACON structure has some disadvantages like the super-network has the burden to simultaneously satisfy all the error constraints by which the number of nodes in the hidden layers tends to be large. The structure of the network is All-Classes-in-One-Network (ACON), shown in Figure 1(a) where one single network is designed to classify all the classes but in One-Class-in-One-Network (OCON), shown in Figure 1(b) a single network is dedicated to recognize one particular class. For each class, a network is created with all the training samples of that class as positive examples, called the class-one, and the negative examples for that class i.e. exemplars from other classes, constitute the class-two. Thus, this classification problem is a two-class partitioning problem. So far, as implementation is concerned, the structure of the network remains the same for all classes and only the weights vary. As the network remains same, weights are kept in separate files and the identification of input image is made on the basis of feature vector and stored weights applied to the network one by one, for all the classes.

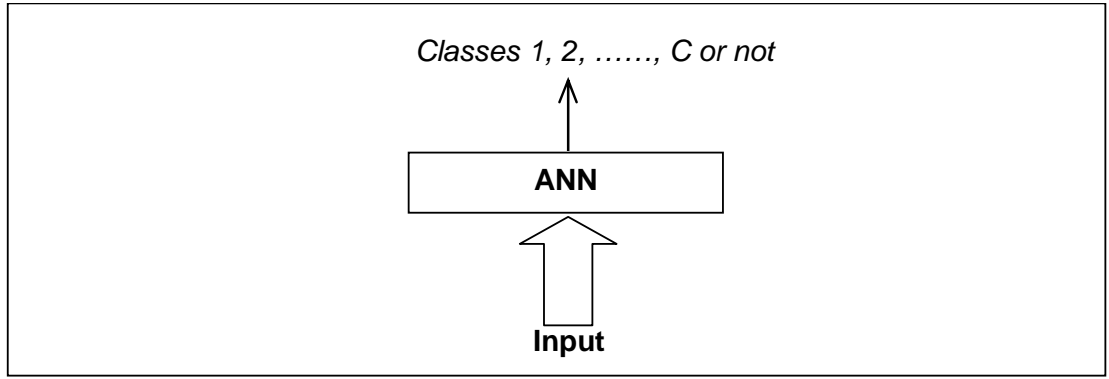

(a)

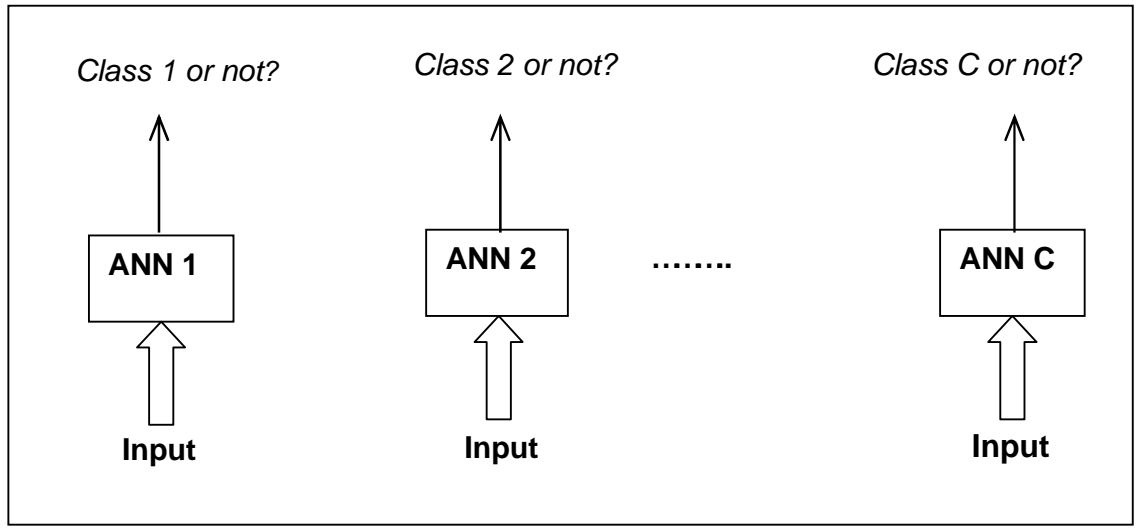

(b)
**Figure 1: a)** All-Classes-in-One-Network (ACON) **b)** One-Class-in-One-Network (OCON).

Empirical results confirm that the convergence rate of ACON degrades drastically with respect to the network size because the training of hidden units is influenced by (potentially conflicting) signals from different teachers. If the topology is changed to One Class in One Network (OCON) structure, where one sub-network is designated and responsible for one class only then each sub-network specializes in distinguishing its own class from the others. So, the number of hidden units is usually small.

**2B. Training of an ANN**
In the training phase the main goal is to utilize the resources as much as possible and speed-up the computation process. Hence, the computation involved in training is distributed over the system to reduce response time. The training procedure can be given as:

(1) Retrieve the topology of the neural network given by the user,
(2) Initialize required parameters and weight vector necessary to train the network,
(3) Train the network as per network topology and available parameters for all exemplars of different classes,
(4) Run the network with test vectors to test the classification ability,
(5) If the result found from step 4 is not satisfactory, loop back to step 2 to change the parameters like learning parameter, momentum, number of iteration or even the weight vector,
(6) If the testing results do not improve by step 5, then go back to step 1,
(7) The best possible (optimal) topology and associated parameters found in step 5 and step 6 are stored.

Although we have parallel systems already in use but some problems cannot exploit advantages of these systems because of their inherent sequential execution characteristics. Therefore, it is necessary to find an equivalent algorithm, which is executable in parallel.

In case of OCON, different individual small networks with least amount of load, which are responsible for different classes (e.g. k classes), can easily be trained in k different processors and the training time must reduce drastically. To fit into this parallel framework previous training procedure can be modified as follows:

(1) Retrieve the topology of the neural network given by the user,
(2) Initialize required parameters and weight vector necessary to train the network,
(3) Distribute all the classes (say k) to available processors (possibly k) by some optimal process allocation algorithm,
(4) Ensure the retrieval the exemplar vectors of respective classes by the corresponding processors,
(5) Train the networks as per network topology and available parameters for all exemplars of different classes,
(6) Run the networks with test vectors to test the classification ability,
(7) If the result found from step 6 is not satisfactory, loop back to step 2 to change the parameters like learning parameter, momentum, number of iteration or even the weight vector,
(8) If the testing results do not improve by step 5, then go back to step 1,
(9) The best possible (optimal) topology and associated parameters found in step 7 and step 8 are stored,
(10) Store weights per class with identification in more than one computer [2].

During the training of two different topologies (OCON and ACON), we used total 200 images of 10 different classes and the images are with different poses and also with different illuminations. Sample images used during training are shown Figure 2. We implemented both the topologies using MATLAB. At the time of training of our systems for both the topologies, we set maximum number of possible epochs (or iterations) to 700000. The training stops if the number of iterations exceeds this limit or performance goal is met. Here, performance goal was considered as $10^{-6}$. We have taken total 10 different training runs for 10 different classes for OCON and one single training run for ACON for 10 different classes. In case of the OCON networks, performance goal was met for all the 10 different training cases, and also in lesser amount of time than ACON. After the completion of training phase of our two different topologies we tested our both the network using the images of testing class which are not used in training.

**2C. Testing Phase**
During testing, the class found in the database with minimum distance does not necessarily stop the testing procedure. Testing is complete after all the registered classes are tested. During testing some points were taken into account, those are:

(1) The weights of different classes already available are again distributed in the available computer to test a particular image given as input,
(2) The allocation of the tasks to different processors is done based on the testing time and inter-processor communication overhead. The communication overhead should be much less than the testing time for the success of the distribution of testing, and
(3) The weight vector of a class matter, not the computer, which has computed it.

The testing of a class can be done in any computer as the topology and the weight vector of that class is known. Thus, the system can be fault tolerant [2]. At the time of testing, we used total 200 images. Among 200 images 100 images are taken from the same classes those are used during the training and 100 images from other classes are not used during the training time. In the both topology (ACON and OCON), we have chosen 20 images for testing, in which 10 images from same class those are used during the training as positive exemplars and other 10 images are chosen from other classes of the database as negative exemplars.

### 2D. Performance measurement
Performance of this system can be measured using following parameters:
(1) resource sharing: Different terminals remain idle most of the time can be used as a part of this system. Once the weights are finalized anyone in the net, even though not satisfying the optimal testing time criterion, can use it. This can be done through Internet attempting to end the "tyranny of geography",
(2) high reliability: Here we will be concerned with the reliability of the proposed system, not the inherent fault tolerant property of the neural network. Reliability comes from the distribution of computed weights over the system. If any of the computer(or processor) connected to the network goes down then the system works Some applications like security monitoring system, crime prevention system require that the system should work, whatever may be the performance,
(3) cost effectiveness: If we use several small personal computers instead of high-end computing machines, we achieve better price/performance ratio,
(4) incremental growth: If the number of classes increases, then the complete computation including the additional complexity can be completed without disturbing the existing system. Based on the analysis of performance of our two different topologies, if we see the recognition rates of OCON and ACON in Table 1 and Table 2 OCON is showing better recognition rate than ACON. Comparison in terms of training time can easily be observed in figures 3 (Figure 3 (a) to (k)). In case of OCON, performance goals met for 10 different classes are $9.99999e-007$, $1e-006$, $9.99999e-007$, $9.99998e-007$, $1e-006$, $9.99998e-007$, $1e-006$, $9.99997e-007$, $9.99999e-007$ respectively, whereas  for ACON it is $0.0100274$. Therefore, it is pretty clear that OCON requires less computational time to finalize a network to use.

## 3. PRINCIPAL COMPONENT ANALYSIS
The Principal Component Analysis (PCA) [5] [6] [7] uses the entire image to generate a set of features in the both network topology OCON and ACON and does not require the location of individual feature points within the image. We have implemented the PCA transform as a reduced feature extractor in our face recognition system. Here, each of the visual face images is projected into the eigenspace created by the eigenvectors of the covariance matrix of all the training images for both the ACON and OCON networks. Here, we have taken the number of eigenvectors in the eigenspace as 40 because eigenvalues for other eigenvectors are negligible in comparison to the largest eigenvalues.

## 4.  EXPERIMENTS RESULTS USING OCON AND ACON
This work has been simulated using MATLAB 7 in a machine of the configuration 2.13GHz Intel Xeon Quad Core Processor and 16 GB of Physical Memory. We have analyzed the performance of our method using YALE B database which is a collection of visual face images with various poses and illumination.

**4A. YALE Face Database B**
This work has been simulated using MATLAB 7 in a machine of the configuration 2.13GHz Intel Xeon Quad Core Processor and 16 GB of Physical Memory. We have analyzed the performance of our method using YALE B database which is a collection of visual face images with various poses and illumination. This database contains 5760 single light source images of 10 subjects each seen under 576 viewing conditions (9 poses x 64 illumination conditions). For every subject in a particular pose, an image with ambient (background) illumination was also captured. Hence, the total number of images is 5850. The total size of the compressed database is about 1GB. The 65 (64 illuminations + 1 ambient) images of a subject in a particular pose have been "tarred" and "gzipped" into a single file. There were 47 (out of 5760) images whose corresponding strobe did not go off. These images basically look like the ambient image of the subject in a particular pose. The images in the database were captured using a purpose-built illumination rig. This rig is fitted with 64 computer controlled strobes. The 64 images of a subject in a particular pose were acquired at camera frame rate (30 frames/second) in about 2 seconds, so there is only small change in head pose and facial expression for those 64 (+1 ambient) images. The image with ambient illumination was captured without a strobe going off. For each subject, images were captured under nine different poses whose relative positions are shown below. Note the pose 0 is the frontal pose. Poses 1, 2, 3, 4, and 5 were about 12 degrees from the camera optical axis (i.e., from Pose 0), while poses 6, 7, and 8 were about 24 degrees. In the Figure 2 sample images of per subject per pose with frontal illumination. Note that the position of a face in an image varies from pose to pose but is fairly constant within the images of a face seen in one of the 9 poses, since the 64 (+1 ambient) images were captured in about 2 seconds. The acquired images are 8-bit (gray scale) captured with a Sony XC-75 camera (with a linear response function) and stored in PGM raw format. The size of each image is 640(w) x 480 (h) [9].

In our experiment, we have chosen total 400 images for our experiment purpose. Among them 200 images are taken for training and other 200 images are taken for testing purpose from 10 different classes. In the experiment we use total two different networks: OCON and ACON. All the recognition results of OCON networks are shown in Table 1, and all the recognition results of ACON network are shown in Table 2. During training, total 10 training runs have been executed for 10 different classes. We have completed total 10 different testing for OCON network using 20 images for each experiment. Out of those 20 images, 10 images are taken form the same classes those were used during training, which acts as positive exemplars and rest 10 images are taken from other classes that acts as negative exemplars for that class. In case of OCON, system achieved 100% recognition rate for all the classes. In case of the ACON network, only one network is used for 10 different classes. During the training we achieved 100% as the highest recognition rate, but like OCON network not for all the classes. For ACON network, on an average, 88% recognition rate was achieved.

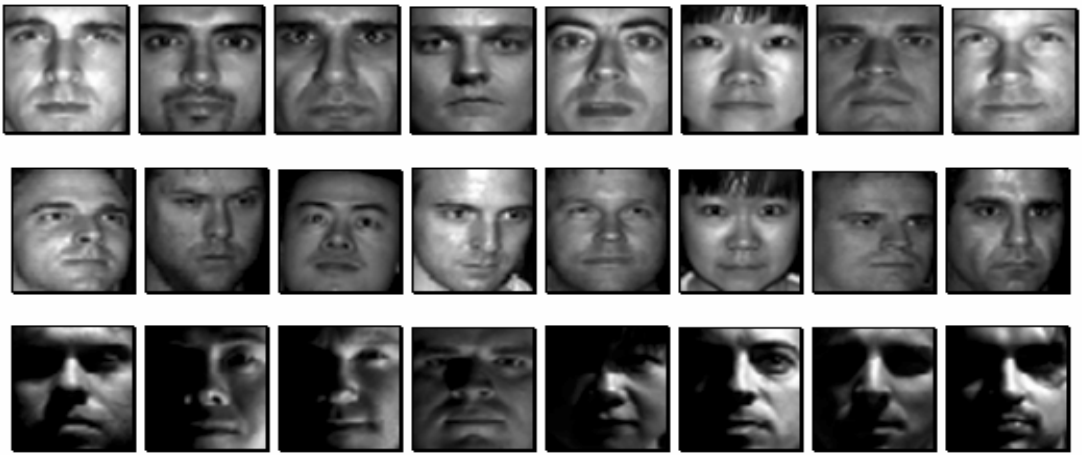

**Figure 2:** Sample images of YALE B database with different Pose and different illumination.

| Class | Total number of testing images | Number of images from the training class | Number of images from other classes | Recognition rate |
|---|---|---|---|---|
| Class-1 | 20 | 10 | 10 | 100% |
| Class-2 | 20 | 10 | 10 | 100% |
| Class-3 | 20 | 10 | 10 | 100% |
| Class-4 | 20 | 10 | 10 | 100% |
| Class-5 | 20 | 10 | 10 | 100% |
| Class-6 | 20 | 10 | 10 | 100% |
| Class-7 | 20 | 10 | 10 | 100% |
| Class-8 | 20 | 10 | 10 | 100% |
| Class-9 | 20 | 10 | 10 | 100% |
| Class-10 | 20 | 10 | 10 | 100% |

**Table 1:** Experiments Results for OCON.

| Class | Total number of testing images | Number of images from the training class | Number of images from other classes | Recognition rate |
|---|---|---|---|---|
| Class - 1 | 20 | 10 | 10 | 100% |
| Class - 2 | 20 | 10 | 10 | 100% |
| Class - 3 | 20 | 10 | 10 | 90% |
| Class - 4 | 20 | 10 | 10 | 80% |
| Class - 5 | 20 | 10 | 10 | 80% |
| Class - 6 | 20 | 10 | 10 | 80% |
| Class - 7 | 20 | 10 | 10 | 90% |
| Class - 8 | 20 | 10 | 10 | 100% |
| Class - 9 | 20 | 10 | 10 | 90% |
| Class-10 | 20 | 10 | 10 | 70% |

**Table 2:** Experiments results for ACON.

In the Figure 3, we have shown all the performance measure and reached goal during 10 different training runs in case of OCON network and also one training phase of ACON network.

We set highest epochs 700000, but during the training, in case of all the OCON networks, performance goal was met before reaching maximum number of epochs. All the learning rates with required epochs of OCON and ACON networks are shown at column two of Table 3.

In case of the OCON network, if we combine all the recognition rates we have the average recognition rate is 100%. But in case of ACON network, 88% is the average recognition rate i.e.

we can say that OCON showing better performance, accuracy and speed than ACON. Figure 4 presents a comparative study on ACON and OCON results.

| Total no. of iterations | Learning Rate (lr) | Class | Figures | Network Used |
|---|---|---|---|---|
| 290556 | lr > $10^{-4}$ | Class – 1 | Figure 3(a) | OCON |
| 248182 | lr = $10^{-4}$ | Class – 2 | Figure 3(b) | |
| 260384 | lr = $10^{-5}$ | Class – 3 | Figure 3(c) | |
| 293279 | lr < $10^{-4}$ | Class - 4 | Figure 3(d) | |
| 275065 | lr = $10^{-4}$ | Class - 5 | Figure 3(e) | |
| 251642 | lr = $10^{-3}$ | Class – 6 | Figure 3(f) | |
| 273819 | lr = $10^{-4}$ | Class – 7 | Figure 3(g) | |
| 263251 | lr < $10^{-3}$ | Class – 8 | Figure 3(h) | |
| 295986 | lr < $10^{-3}$ | Class – 9 | Figure 3(i) | |
| 257019 | lr > $10^{-6}$ | Class - 10 | Figure 3(j) | |
| Highest epoch reached (7, 00, 000) | Performance goal not met | For all Classes (class -1,…,10) | Figure 3(k) | ACON |

**Table 3:** Learning Rate vs. Required Epochs for OCON and ACON.

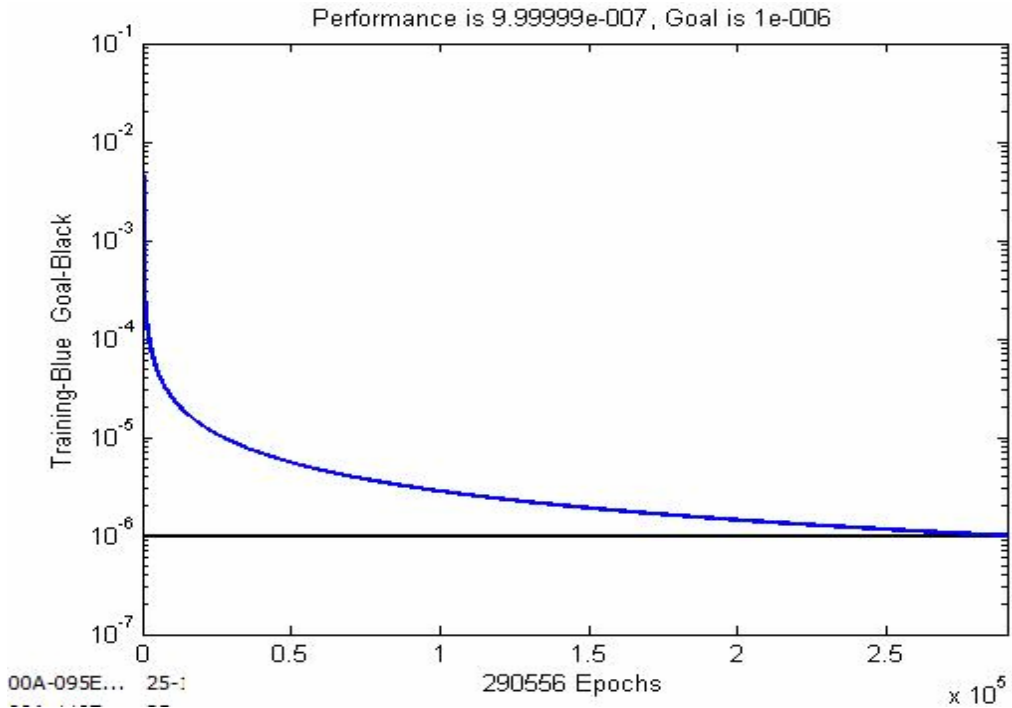

**Figure 3 (a)** Class – 1 of OCON Network.

D. Bhattacharjee, M. K. Bhowmik, M. Nasipuri, D. K. Basu & M. Kundu

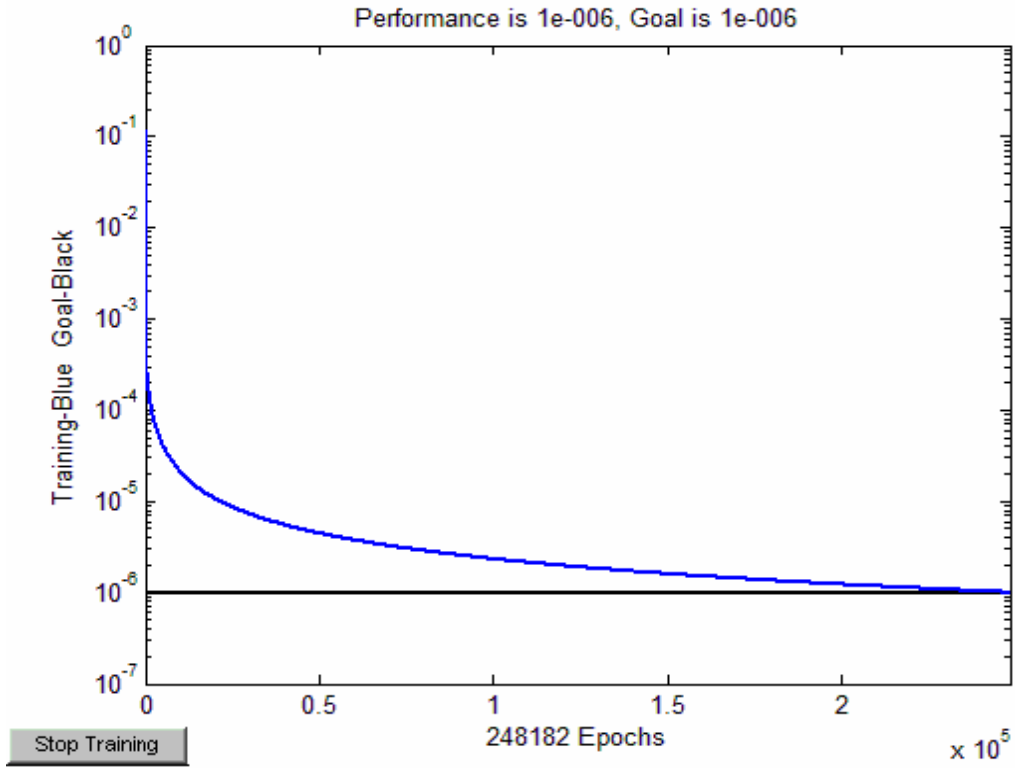

**Figure 3 (b)** Class – 2 of OCON Network.

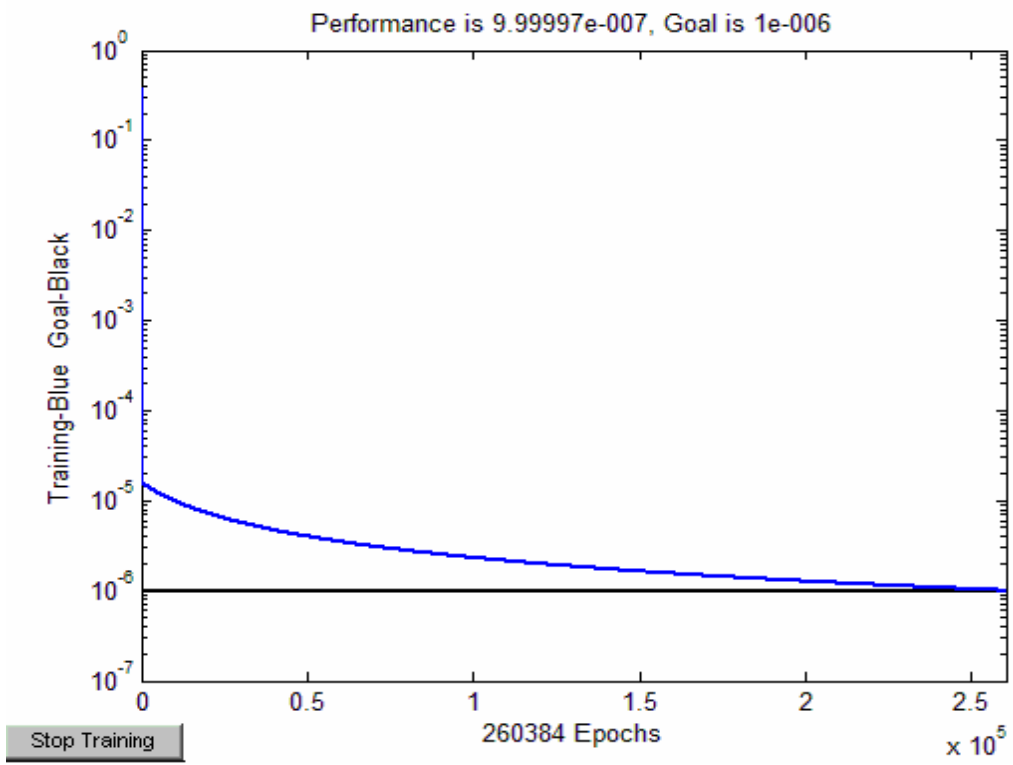

**Figure 3 (c)** Class – 3 of OCON Network.

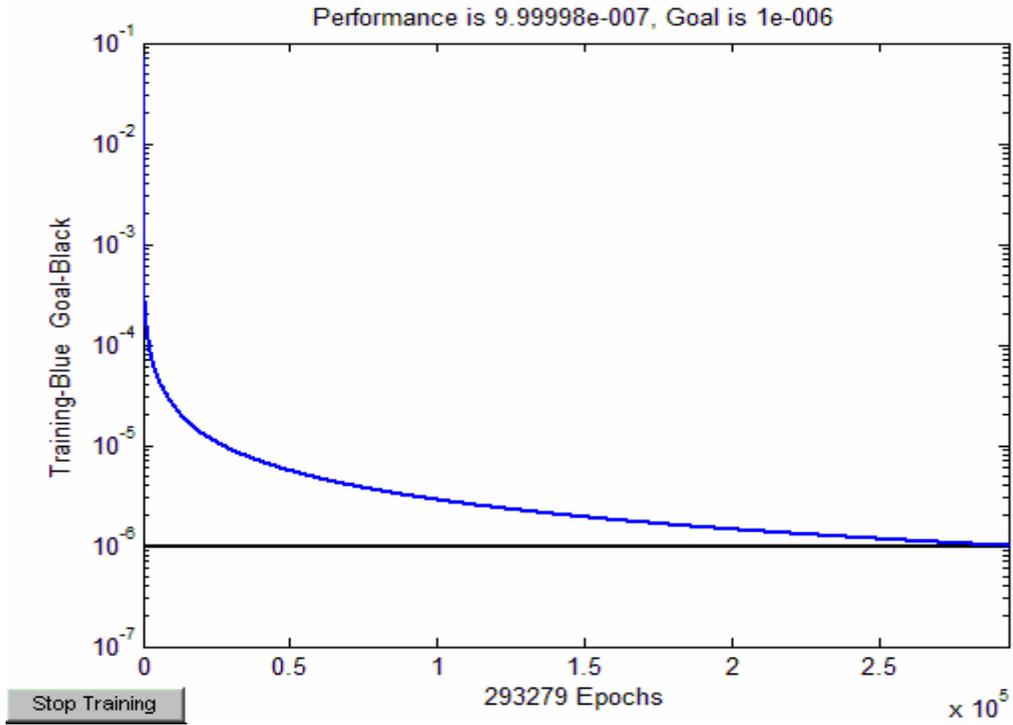

**Figure 3 (d)** Class – 4 of OCON Network.

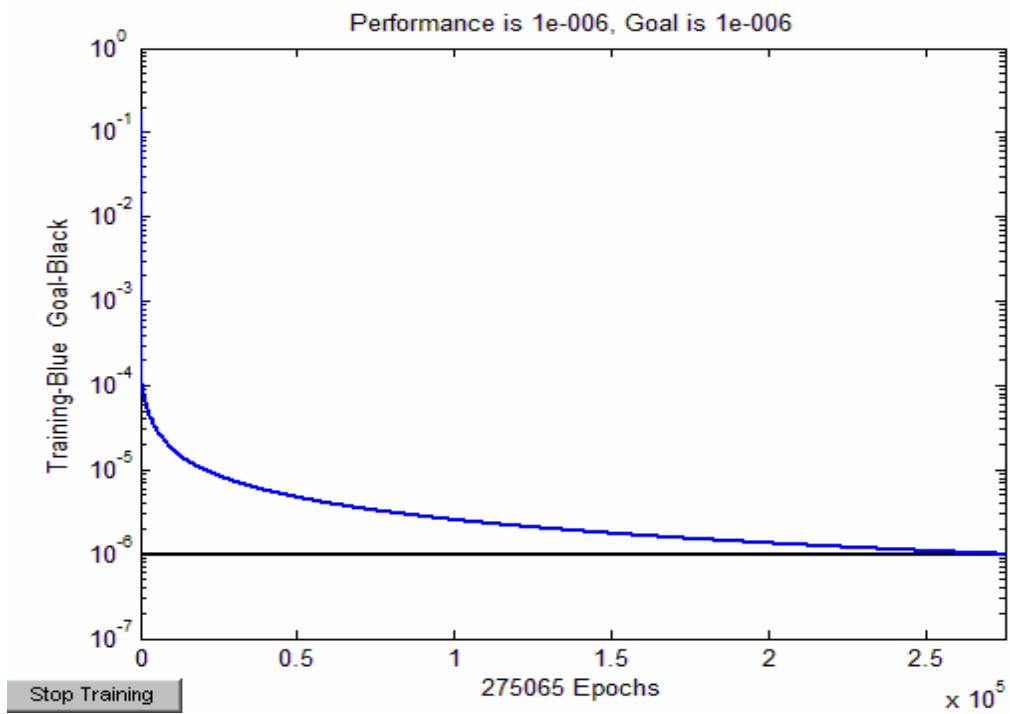

**Figure 3 (e)** Class – 5 of Ocon Network.

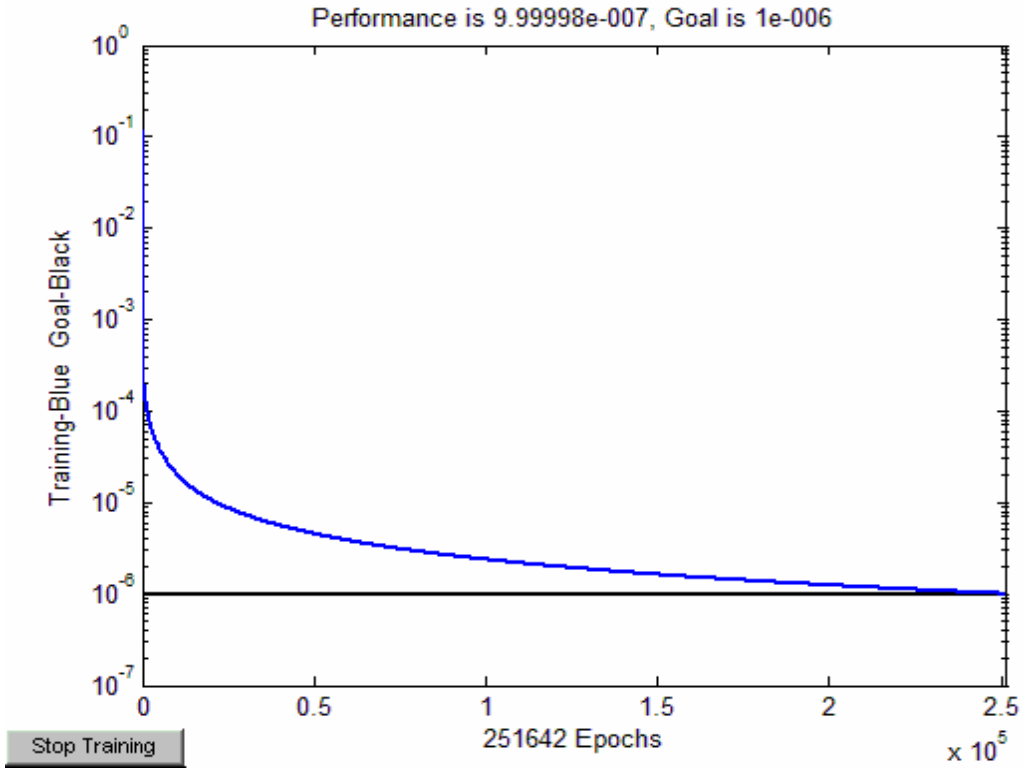

**Figure 3 (f)** Class – 6 of OCON Network.

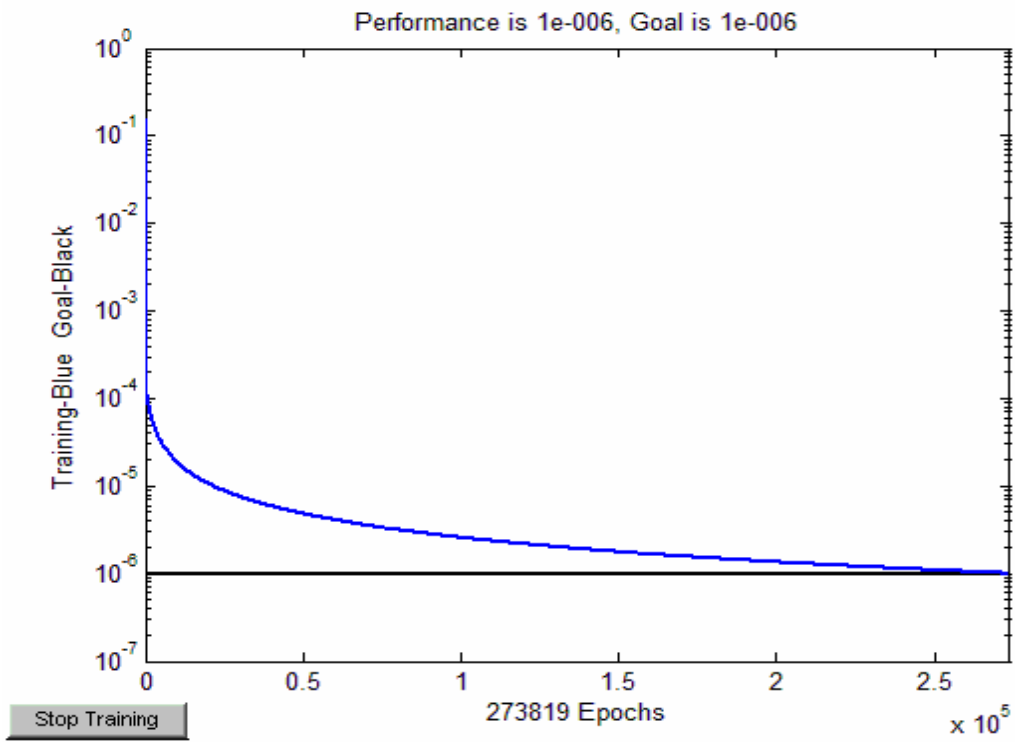

**Figure 3 (g)** Class – 7 of OCON Network.

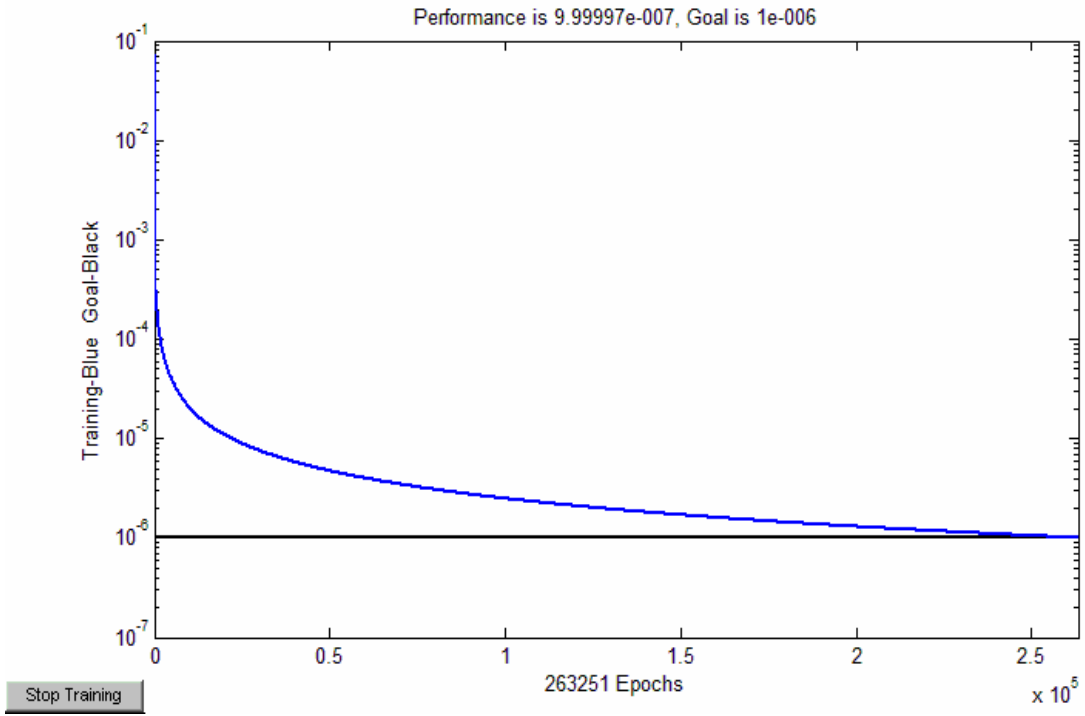

**Figure 3 (h)** Class – 8 of OCON Network.

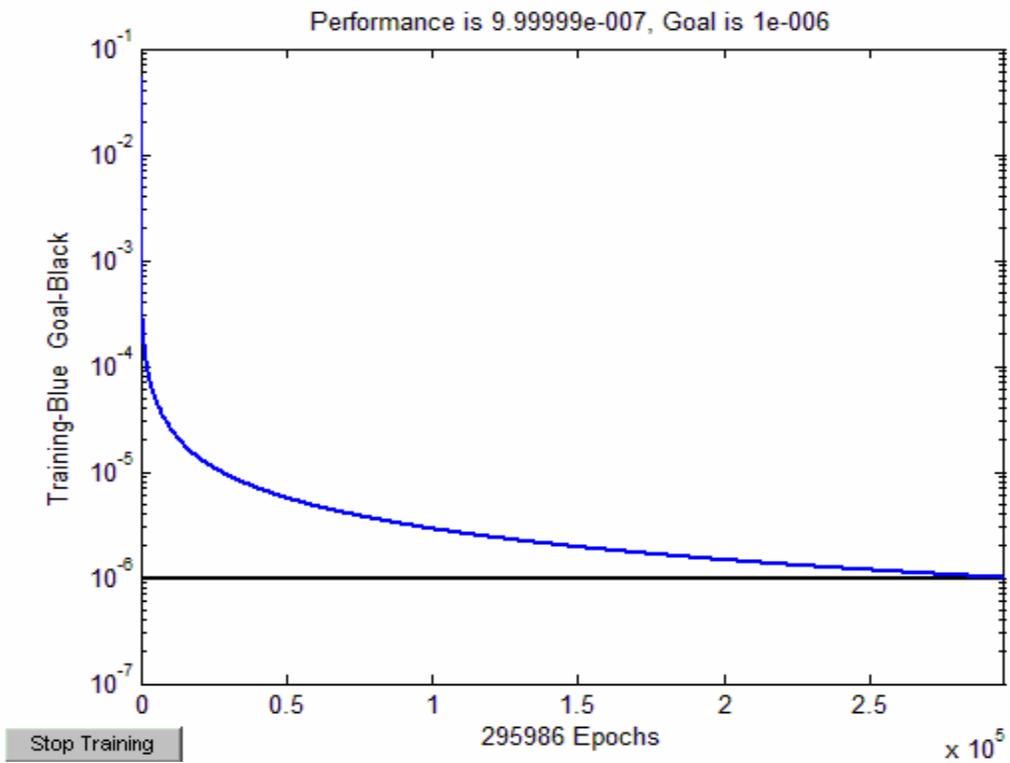

**Figure 3 (i)** Class – 9 of OCON Network.

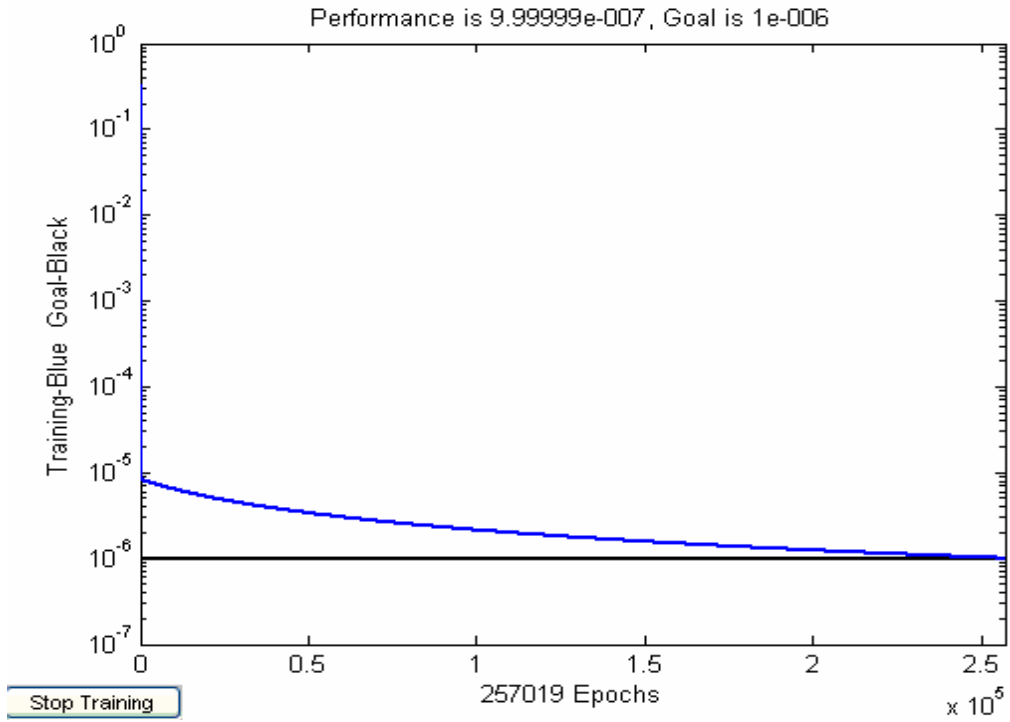

**Figure 3 (j)** Class – 10 of OCON Network.

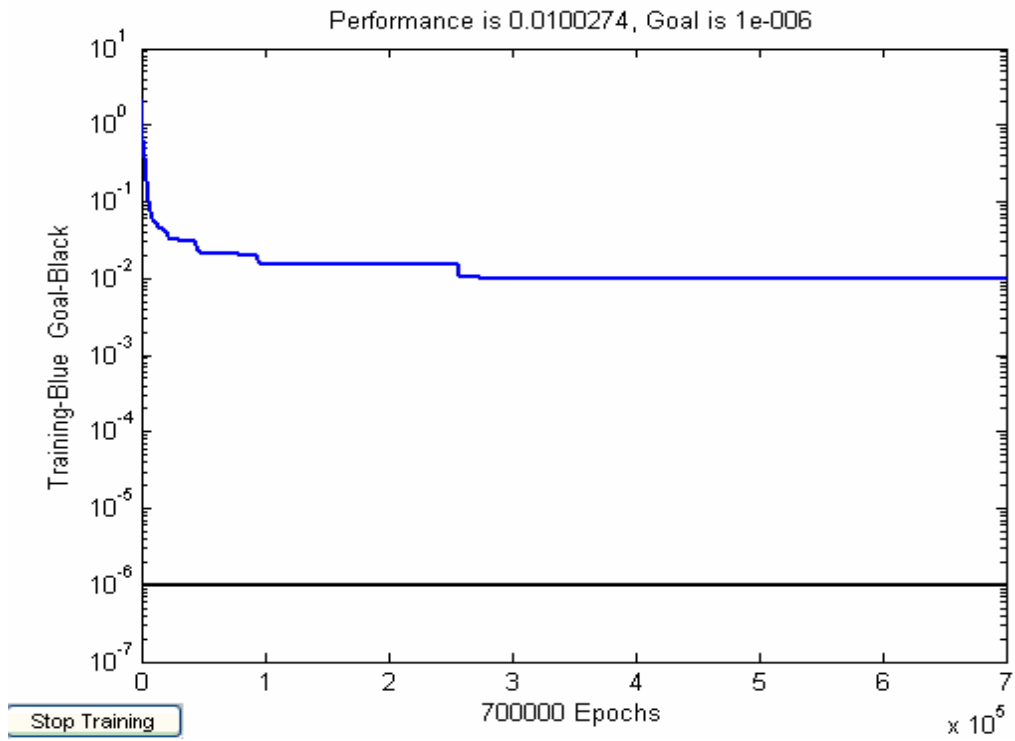

**3 (k)** of ACON Network for all the classes.

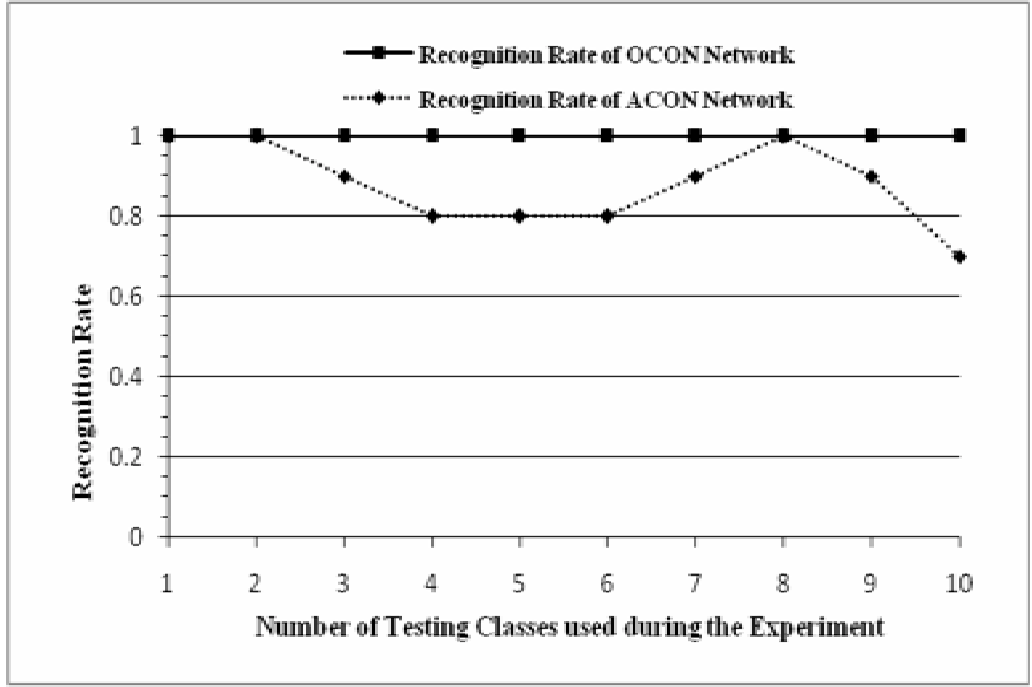

**Figure 4:** Graphical Representation of all Recognition Rate using OCON and ACON Network.

The OCON is an obvious choice in terms of speed-up and resource utilization. The OCON structure of neural network makes it most suitable for incremental training, i.e., network upgrading upon adding/removing memberships. One may argue that compared to ACON structure, the OCON structure is slow in retrieving time when the number of classes is very large. This is not true because, as the number of classes increases, the number of hidden neurons in the ACON structure also tends to be very large. Therefore ACON is slow. Since the computation time of both OCON and ACON increases as number of classes grows, a linear increasing of computation time is expected in case of OCON, which might be exponential in case of ACON.

## 5. CONCLUSION
In this paper, two general architectures for a Multilayer Perceptron (MLP) have been demonstrated. The first architecture is All-Class-in-One-Network (ACON) where all the classes are placed in a single network and the second one is One-Class-in-One-Network (OCON) where an individual single network is responsible for each and every class. Capabilities of these two architectures were compared and verified in solving human face recognition, which is a complex pattern recognition task where several factors affect the recognition performance like pose variations, facial expression changes, occlusions, and most importantly illumination changes. Both the structures were implemented and tested for face recognition purpose and experimental results show that the OCON structure performs better than the generally used ACON ones in term of training convergence speed of the network. Moreover, the inherent non-parallel nature of ACON has compelled us to use OCON for the complex pattern recognition task like human face recognition.

## ACKNOWLEDGEMENT
Second author is thankful to the project entitled "Development of Techniques for Human Face Based Online Authentication System Phase-I" sponsored by Department of Information Technology under the Ministry of Communications and Information Technology, New Delhi-110003, Government of India Vide No. 12(14)/08-ESD, Dated 27/01/2009 at the Department of Computer Science & Engineering, Tripura University-799130, Tripura (West), India for providing the necessary infrastructural facilities for carrying out this work. The second author is also grateful to the Honorable Vice Chancellor Prof. A. Saha and Prof. B.K. De, Dean of Science of Tripura University (A Central University), Tripura, India for their kind support to carry out this research work.

## REFERENCES

1. M. K. Bhowmik, "Artificial Neural Network as a Soft Computing Tool – A case study", In Proceedings of National Seminar on Fuzzy Math. & its application, Tripura University, November 25 – 26, 2006, pp: 31 – 46.

2. M. K. Bhowmik, D. Bhattacharjee and M. Nasipuri, "Topological Change in Artificial Neural Network for Human Face Recognition", In Proceedings of National Seminar on Recent Development in Mathematics and its Application, Tripura University, November 14 – 15, 2008, pp: 43 – 49.

3. A. Abraham, "Artificial Neural Network", Oklahoma Tate University, Stillwater, OK, USA.

4. http://www.acm.org/ubiquity/homepage.html.

5. I. Aleksander and H. Morton, "An introduction to Neural Computing," Chapman & Hall, London, 1990.

6. R. Hecht-Nielsen, "Neurocomputing," Addison-Wesley, 1990.

7. M. Turk and A. Pentland, "Eigenfaces for recognition", Journal of Cognitive Neuro-science, March 1991. Vol. 3, No-1, pp. 71-86.

8. L. Sirovich and M. Kirby, "A low-dimensional procedure for the characterization of human faces," J. Opt. Soc. Amer. A 4(3), pp. 519-524, 1987.

9. A. S. Georghiades, P. N. Belhumeur and D. J. Kriegnab, "From Few to Many: Illumination Cone Models for face Recognition under Variable Lighting and Pose", IEEE Trans. Pattern Anal. Mach. Intelligence, 2001, vol. 23, No. 6, pp. 643 – 660.

10. D. Bhattacharjee, "Exploiting the potential of computer network to implement neural network in solving complex problem like human face recognition," Proc. of national Conference on Networking of Machines, Microprocessors, IT, and HRD-need of the nation in the next millennium, Kalyani Engg. College, Kalyani, West Bengal, 1999.

11. Pradeep K. Sinha, "Distributed Operating Systems-Concepts and Design," PHI, 1998. [8] M. K. Bhowmik, D. Bhattacharjee, M. Nasipuri, D. K. Basu and M. Kundu; "Classification of Fused Face Images using Multilayer Perceptron Neural Network", proceeding of International Conference on Rough sets, Fuzzy sets and Soft Computing, Nov 5–7, 2009, organized by Department of Mathematics, Tripura University pp. 289-300.

12. M.K. Bhowmik, D. Bhattacharjee, M. Nasipuri, D.K. Basu and M. Kundu, "Classification of Polar-Thermal Eigenfaces using Multilayer Perceptron for Human Face Recognition", proceedings of the 3$^{rd}$ IEEE Conference on Industrial and Information Systems (ICIIS-2008), IIT Kharagpur, India, Dec 8-10, 2008, pp. 118.

13. M.K. Bhowmik, D. Bhattacharjee, M. Nasipuri, D.K. Basu and M. Kundu, "Classification of Log-Polar-Visual Eigenfaces using Multilayer Perceptron for Human Face Recognition", proceedings of The 2$^{nd}$ International Conference on Soft computing (ICSC-2008), IET, Alwar, Rajasthan, India, Nov 8–10, 2008, pp.107-123.

14. M.K. Bhowmik, D. Bhattacharjee, M. Nasipuri, D.K. Basu and M. Kundu, "Human Face Recognition using Line Features", proceedings of National Seminar on Recent Advances on Information Technology (RAIT-2009), Indian School of Mines University, Dhanbad, Feb 6-7,2009, pp. 385-392.

15. P. Raviram, R.S.D. Wahidabanu Implementation of artificial neural network in concurrency control of computer integrated manufacturing (CIM) database, International Journal of Computer Science and Security (IJCSS), Volume 2, Issue 5, pp. 23-25, September/October 2008.

16. Teddy Mantoro, Media A. Ayu, "Toward The Recognition Of User Activity Based On User Location In Ubiquitous Computing Environments," International Journal of Computer Science and Security (IJCSS)Volume 2, Issue 3, pp. 1-17, May/June 2008.

17. Sambasiva Rao Baragada, S. Ramakrishna, M.S. Rao, S. Purushothaman , "Implementation of Radial Basis Function Neural Network for Image Steganalysis," International Journal of Computer Science and Security (IJCSS) Volume 2, Issue 1, pp. 12-22, January/February 2008.